\documentclass[11pt]{article}

\usepackage[preprint]{acl}
\usepackage{enumitem}
\usepackage{booktabs}
\usepackage{multirow}
\usepackage[table]{xcolor}
\usepackage{url}

\usepackage{array}
\usepackage{graphicx}
\usepackage{times}
\usepackage{adjustbox}

\usepackage{latexsym}
\usepackage{amsmath}
\usepackage{amssymb}
\usepackage{booktabs}
\usepackage{graphicx}
\usepackage{multirow}
\usepackage[table]{xcolor}

\usepackage[T1]{fontenc}
\usepackage[table]{xcolor}
\usepackage{booktabs}
\usepackage{multirow}
\usepackage{array}
\usepackage{tabularx}
\usepackage{pifont}
\usepackage[most]{tcolorbox}

\usepackage[most]{tcolorbox}

\tcbset{
  casebox/.style={
    enhanced,
    breakable,
    width=\linewidth,
    boxrule=0.6pt,
    arc=1.5mm,
    left=1.5mm,
    right=1.5mm,
    top=1mm,
    bottom=1mm,
    fonttitle=\bfseries,
    before skip=6pt,
    after skip=6pt
  }
}

\newtcolorbox{questionbox}[1][]{
  casebox,
  title={Question},
  colback=gray!5,
  colframe=gray!70!black,
  #1
}

\newtcolorbox{standardbox}[1][]{
  casebox,
  title={Standard Response},
  colback=green!4,
  colframe=green!45!black,
  #1
}

\newtcolorbox{rolloutonebox}[1][]{
  casebox,
  title={Model Sampling Rollout 1},
  colback=blue!3,
  colframe=blue!55!black,
  #1
}

\newtcolorbox{rollouttwobox}[1][]{
  casebox,
  title={Model Sampling Rollout 2},
  colback=orange!5,
  colframe=orange!75!black,
  #1
}

\newtcolorbox{rolloutthreebox}[1][]{
  casebox,
  title={Model Sampling Rollout 3},
  colback=red!4,
  colframe=red!65!black,
  #1
}
\usepackage{listings}
\usepackage{xcolor}

\usepackage[utf8]{inputenc}

\usepackage{microtype}

\usepackage{inconsolata}

\usepackage{graphicx}

\title{RASFT: Rollout-Adaptive Supervised Fine-Tuning for Reasoning
}

\author{
\textbf{Yongliang Miao}\textsuperscript{1,*},
\textbf{Fengyuan Liu}\textsuperscript{1,*},
\textbf{Wei Shi}\textsuperscript{2}\\
\textbf{Yanguang Liu}\textsuperscript{3},
\textbf{Fei Sun}\textsuperscript{4},
\textbf{Na Zou}\textsuperscript{2},
\textbf{Mengnan Du}\textsuperscript{1,\textdagger}\\
\textsuperscript{1}The Chinese University of Hong Kong, Shenzhen \,
\textsuperscript{2}Shanghai Artificial Intelligence Laboratory \,\\
\textsuperscript{3}New Jersey Institute of Technology \,
\textsuperscript{4}Institute of Computing Technology, CAS\\
\small\texttt{r130026108@gmail.com, mengnandu@cuhk.edu.cn}\\
\small\textsuperscript{*}Equal contribution.
\qquad
\small\textsuperscript{\textdagger}Corresponding author.
}

\begin{document}
\maketitle
\begin{abstract}
Supervised fine-tuning (SFT) is a prevailing method for adapting large language models to reasoning tasks by imitating offline expert demonstrations, often treating a single expert trajectory as the target behavior. However, reasoning is not simple path imitation: rigidly following one demonstrated solution may overfit to surface forms and suppress the model's own reasoning distribution. We propose \textbf{Rollout-Adaptive Supervised Fine-Tuning (RASFT)}, a policy-aware SFT framework that calibrates expert supervision according to problem-level solvability estimated from verified on-policy rollouts. For each problem, RASFT strengthens expert guidance when the current policy struggles, while relaxing rigid imitation and incorporating correct self-generated trajectories when the model already exhibits reliable reasoning behavior. To preserve useful reasoning priors, RASFT further introduces a clipped inverse ratio between the frozen reference model and the current policy to constrain excessive policy drift. Experiments across multiple models on six mathematical reasoning benchmarks and two code reasoning benchmarks show that RASFT achieves better overall performance than SFT, SFT variants, and representative RL methods. The code is available at \url{https://github.com/zjd1sq/RASFT}.
\end{abstract}

\section{Introduction}

Supervised fine-tuning (SFT) has been widely used to adapt large language models (LLMs) to reasoning domains~\cite{ouyang2022training,chung2022scaling,mukherjee2023orca,yue2023mammoth}.
Standard SFT relies on offline expert demonstrations and implicitly assumes that expert trajectories are the targets to be imitated~\cite{ouyang2022training,wang2023selfinstruct,zhou2023lima}. However, reasoning is not simple imitation: one problem can often be solved through multiple valid paths, while an offline expert response presents only one possible reasoning trajectory~\cite{wei2022chain,zelikman2022star,yuan2023scaling}.
Meanwhile, pretrained LLMs may already possess rich potential reasoning distributions~\cite{zelikman2022star,mukherjee2023orca}.
If fine-tuning overly follows a single expert trajectory, the model may fit the surface form of the demonstration, thereby weakening its original reasoning distribution and limiting the activation of its own reasoning ability~\cite{chu2025sft,mukherjee2023orca}.

This limitation has motivated recent SFT variants to reconsider how expert demonstrations should be optimized. 
DFT~\cite{wu2025generalization} weakens uniform token-level imitation by rescaling the learning signal according to the model's confidence. 
ASFT~\cite{zhu2025anchored} further anchors this optimization to a reference model to improve stability and reduce distributional drift. 
ProFiT~\cite{liu2026profit} reduces unnecessary fitting by emphasizing high-value supervision signals within the expert response. 
These methods make SFT less blindly imitative at the token or objective level, but their adaptivity remains largely demonstration-internal, not policy-aware: the expert trajectory is still optimized as the default target, leaving the objective unable to distinguish when expert imitation provides necessary correction and when it becomes rigid path fitting for the current policy.

This motivates a policy-adaptive view of SFT in reasoning domains: \textit{expert demonstrations should serve as problem-level guidance whose strength is calibrated by the current policy's ability to solve each problem, rather than only by signals within the offline trajectory.}
The challenge is therefore how to determine the role of expert supervision for each problem under the current policy.
When the model still struggles, expert trajectories should provide stronger corrective guidance; when the model already exhibits reliable reasoning behavior, rigid imitation of the same expert path becomes less necessary. 
Therefore, effective SFT should move beyond static demonstration fitting and adjust expert supervision according to policy-dependent problem difficulty, while avoiding unnecessary suppression of the model's own reasoning distribution.

To this end, we propose \textbf{Rollout-Adaptive Supervised Fine-Tuning (RASFT)}, a policy-aware SFT framework that adapts expert supervision using the current model's rollout behavior.
 For each problem, RASFT constructs a local candidate pool consisting of the offline expert trajectory and verified correct trajectories generated by the current model. It estimates a problem-level solvability score from the success rate of model-generated rollouts, which reflects how well the current policy can solve the problem. Based on this score, RASFT increases the influence of expert demonstrations on difficult problems and attenuates it when the model already exhibits reliable reasoning behavior. Furthermore, to conservatively activate the model's reasoning ability rather than overwrite it through excessive imitation, RASFT introduces an inverse ratio between the frozen reference model and the current policy to constrain excessive policy drift.

Empirically, RASFT consistently outperforms SFT-style baselines across mathematical and code reasoning tasks.
It achieves a $10.9\%$ relative gain on Qwen2.5-Math-1.5B math reasoning ($25.00 \rightarrow 27.72$) and up to $26.9\%$ on Llama-3.2-3B code generation ($24.93 \rightarrow 31.63$), suggesting that rollout-adaptive supervision brings benefits beyond stronger demonstration fitting.
Compared with GRPO~\cite{shao2024deepseekmath}, RASFT further improves average math accuracy by $15.9\%$ ($20.25 \rightarrow 23.47$), highlighting the robustness of combining expert guidance with rollout-based adaptation.

\noindent Our key contributions are summarized as follows:

\begin{itemize}[leftmargin=*, topsep=2pt, itemsep=1pt, parsep=0pt, partopsep=0pt]
    \item We identify a key limitation of existing SFT variants for reasoning: their adaptivity mainly operates within offline expert demonstrations, rather than calibrating expert supervision according to the current policy's problem-level ability.

    \item We propose RASFT, a rollout-adaptive SFT framework that uses verified on-policy rollouts to estimate problem solvability and dynamically balance expert guidance with self-generated correct reasoning trajectories.

     \item Extensive experiments on mathematical and code reasoning tasks show that RASFT outperforms SFT-style baselines across multiple models, with ablations and RL comparisons further confirming its effectiveness and robustness.

\end{itemize}

\section{Methodology}
\begin{figure*}[t]
  \centering
  \resizebox{0.92\linewidth}{!}{
    \includegraphics{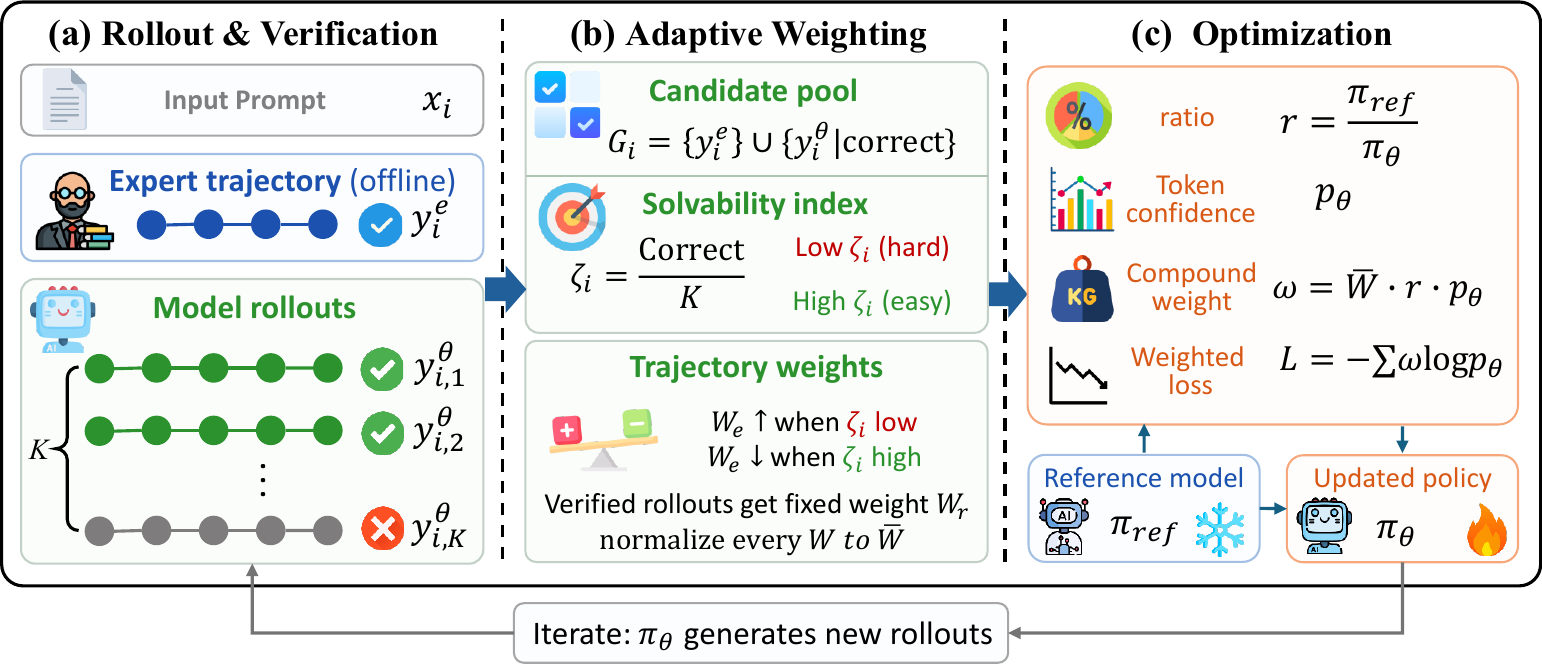}
  }
\caption{
\textbf{RASFT pipeline.}
(a) For each prompt, policy model $\pi_{\theta}$ samples multiple rollouts, which are verified and combined with offline expert trajectory.
(b) Rollout-based solvability  $\zeta_i$, which adaptively calibrates expert and rollout trajectory weights.
(c) RASFT updates the policy model $\pi_{\theta}$ by optimizing candidate trajectories with a compound weight that combines normalized trajectory weights, an inverse policy ratio, and token probability.
}

  \label{fig:method}
  \vspace{-9pt}
\end{figure*}

\subsection{Preliminaries}

We consider supervised post-training for reasoning tasks. 
Let $\mathcal{D}_{\mathrm{off}}=\{(x_i,y_i^{e})\}_{i=1}^{N}$ denote an offline reasoning dataset, where $x_i$ is a problem prompt and $y_i^{e}$ is an expert-written reasoning trajectory. 
A language model parameterized by $\theta$ defines an autoregressive policy
\begin{equation}
\pi_{\theta}(y \mid x)
=
\prod_{t=1}^{|y|}
\pi_{\theta}(y_t \mid x,y_{<t}).
\end{equation}
Conventional supervised fine-tuning optimizes the negative log-likelihood of the expert response:
\begin{equation*}
\mathcal{L}_{\mathrm{SFT}}(\theta)
{=}
-\!
\mathop{\mathbb{E}}\limits_{(x,y^{e})\sim \mathcal{D}_{\mathrm{off}}}\!\!
\left[
\sum_{t=1}^{|y^{e}|}
\!\log \pi_{\theta}(y_t^{e}{\mid} x,y_{<t}^{e})\!
\right].
\end{equation*}
This objective treats each expert trajectory as the desired target behavior and applies uniform token-level imitation along the given response.

However, reasoning differs from ordinary sequence imitation: the expert trajectory $y^{e}$ provides only one possible solution path, while the model generates according to its own policy $\pi_{\theta}(\cdot\mid x)$. We aim to use expert responses as supervision with less excessive imitation of trajectories that are redundant, misaligned, or already well captured by the current policy (Figure~\ref{fig:method}).

\subsection{RASFT Framework}

\noindent\textbf{Policy-aware regularization.} RASFT optimizes expert and model-generated reasoning trajectories using policy-aware regularization and trajectory-level adaptive weighting, moving away from treating offline expert responses as uniformly imitative targets. Instead of framing post-training through an explicit reinforcement learning lens with rigid reward or advantage formulations, we view expert trajectories and self-generated successful behaviors as an aligned candidate pool where optimization scales dynamically according to policy-relative properties.

The core motivation is that large language models already possess substantial reasoning capacity from pretraining. Thus, the objective of post-training is not to force the mechanical memorization of static demonstrations, but to leverage expert knowledge to activate and refine the model's latent policy distribution. Let $\pi_{\mathrm{ref}}$ denote the frozen initial reference model, and let $\pi_{\theta}$ denote the current trainable policy. For an arbitrary target trajectory $y_{i,j}=(y_{i,j,1},\ldots,y_{i,j,|y_{i,j}|})$ associated with prompt $x_i$, where $j$ indexes a candidate trajectory, the target-token probability under the active policy is defined as:
\begin{equation*}
p_{i,j,t}^{\theta} =\pi_{\theta}(y_{i,j,t}\mid x_i,y_{i,j,<t}).
\end{equation*}

To regularize how aggressively a candidate trajectory updates the policy, RASFT defines a sequence-level inverse importance sampling ratio:
\begin{equation*}
\begin{split}
r_{i,j}(\theta)
&{=}
\left(
\frac{\pi_{\mathrm{ref}}(y_{i,j}\mid x_i)}
{\pi_{\theta}(y_{i,j}\mid x_i)}
\right)^{\frac{1}{|y_{i,j}|}} \\
&{=}
\exp\!\!
\left(\!
\frac{1}{|y_{i,j}|}\!\!
\sum_{t=1}^{|y_{i,j}|}\!
\log\!
\frac{
\pi_{\mathrm{ref}}(y_{i,j,t}{\mid} x_i,y_{i,j,<t})
}{
\pi_{\theta}(y_{i,j,t}{\mid} x_i,y_{i,j,<t})
}\!\!
\right).
\end{split}
\end{equation*}
Here, the numerator reflects the reference model's support for the candidate trajectory, ensuring that updates consistent with the pretrained prior are preserved to prevent distributional drift. Conversely, as $\pi_{\theta}(y_{i,j}\mid x_i)$ grows relative to $\pi_{\mathrm{ref}}(y_{i,j}\mid x_i)$, indicating that the current policy has already shifted toward this behavior, the inverse ratio diminishes the gradient magnitude. This ratio stabilizes optimization against length variations and helps prevent over-imitation of trajectories.

\noindent\textbf{Rollout-based solvability.} To augment offline guidance with online policy feedback, RASFT incorporates self-sampled rollouts. For a given prompt $x_i$, the current model generates $K$ rollouts, which are verified by a task-specific verifier (e.g., final-answer matching). Let $c_{i,k}\in\{0,1\}$ indicate whether the $k$-th rollout is correct. We define the problem-level solvability index as:
\[
\zeta_i = \frac{1}{K} \sum_{k=1}^{K} c_{i,k}.
\]
Crucially, all incorrect rollouts are strictly filtered out from the optimization targets to avoid training noise; they are exclusively utilized to compute $\zeta_i$ as an explicit proxy for problem difficulty. 

For compute efficiency, RASFT does not perform rollouts for every training instance. Instead, it maintains a sliding window over the current expert-sequence loss and allocates rollouts only to instances within this window. Samples outside the window fall back to expert data-only supervision, since very easy instances provide limited exploration value and difficult ones   yield relatively few reliable successful rollouts.

For samples within the window, we construct a local candidate pool $\mathcal{G}_i$ consisting of the expert trajectory and the model's own successful rollouts:
\[
\mathcal{G}_i = \{y_i^{e}\} \cup \{y_{i,k}^{\theta} \mid c_{i,k} = 1\}.
\]
\noindent\textbf{Adaptive weighting and objective.} We assign trajectory-level weights $W_i$ within this pool. For the expert trajectory, we assign a difficulty-adaptive weight using the rollout success rate:
\[
W_i(y_i^e)
=
W_e^{\min}
+
(1-\zeta_i)(W_e^{\max}-W_e^{\min}),
\label{eq:expert-weight}
\]
where $W_e^{\min}$ and $W_e^{\max}$ denote the minimum and maximum expert weights. When $\zeta_i$ is low, the expert trajectory provides stronger corrective supervision; when $\zeta_i$ is high, its influence is weakened to prevent rigid fitting. Self-generated correct rollouts receive a fixed weight $W_i(y_{i,k}^\theta) = W_r$. We then perform a normalization over the candidate pool to obtain the relative trajectory coefficient $\bar{W}_{i,j}$:
\[
\bar{W}_{i,j} = \frac{W_i(y_{i,j})}{\sum_{y_{i,r} \in \mathcal{G}_i} W_i(y_{i,r})}.
\]

Following token-scaled optimization principles, we combine this trajectory coefficient with a token-level confidence factor $p_{i,j,t}^{\theta}$ and the clipped inverse sequence ratio to derive the final compound weight:
\begin{equation}
\omega_{i,j,t}
{=}
\bar{W}_{i,j}
\operatorname{Clip}\!
\left(
r_{i,j}(\theta),
1{-}\epsilon_{\mathrm{low}},
1{+}\epsilon_{\mathrm{high}}
\right)
p_{i,j,t}^{\theta},
\label{eq:weight}
\end{equation}
where $\epsilon_{\mathrm{low}}$ and $\epsilon_{\mathrm{high}}$ bound the policy deviation limits. 
$\bar{W}_{i,j}$, the clipped ratio term, and the token confidence factor are treated as stop-gradient weights. 
The final RASFT objective is minimized as a regularized, weighted sequence-to-sequence loss:
\[
\mathcal{L}(\theta)
=
-
\mathbb{E}_{i}
\left[
\sum_{y_{i,j}\in\mathcal{G}_i}
\sum_{t=1}^{|y_{i,j}|}
\omega_{i,j,t}
\log p_{i,j,t}^{\theta}
\right].
\]

This objective keeps RASFT within a supervised fine-tuning framework, but makes the supervision policy-adaptive rather than uniformly imitative. 
The rollout-derived solvability controls the expert--rollout balance, while the inverse ratio constrains excessive drift from the pretrained policy.

\begin{table*}[t]
    \centering
    \begingroup
    \setlength{\tabcolsep}{3.2pt}
    \renewcommand{\arraystretch}{1.06}

    \newcommand{\method}[1]{#1}
    \newcommand{\model}[1]{\textbf{\textit{#1}}}
    \begin{adjustbox}{max width=\textwidth}
    \begin{tabular}{@{}l*{7}
    {c}@{\hspace{10pt}}*{3}{c}@{}}
        \toprule
        \multirow{2.5}{*}{\textbf{Methods}}
        & \multicolumn{7}{c}{\textbf{Math Benchmarks}}
        & \multicolumn{3}{c}{\textbf{Code Benchmarks}} \\
        \cmidrule(lr){2-8} \cmidrule(lr){9-11}
        & \textbf{Math} & \textbf{Minerva} & \textbf{Olympiad}
        & \textbf{AIME24} & \textbf{AIME25} & \textbf{AMC23} & \textbf{Avg.}
        & \textbf{HumanEval} & \textbf{MBPP} & \textbf{Avg.} \\
        \midrule

        & \multicolumn{7}{l}{\model{Qwen2.5-Math-1.5B}}
        & \multicolumn{3}{l}{\model{Qwen2.5-Coder-3B}} \\

        \method{Base}
        & 23.21 & 5.82 & 13.24 & 1.87 & 1.46 & 18.28 & 10.65
        & 19.55 & 33.40 & 26.48 \\

        \method{SFT}
        & 40.69 & 11.27 & 10.28 & 1.66 & 0.41 & 16.56 & 13.48
        & 55.87 & 48.39 & 52.13 \\

        \method{DFT~\cite{wu2025generalization}}
        & 60.58 & 20.41 & 22.42 & 4.79 & 3.32 & 33.59 & 24.19
        & 63.15 & 56.09 & 59.62 \\

        \method{ASFT~\cite{zhu2025anchored}}
        & 60.89 & 20.18 & 24.11 & 4.37 & \textbf{4.37} & 36.09 & 25.00
        & 60.33 & 47.98 & 54.16 \\

        \method{ProFiT~\cite{liu2026profit}}
        & 58.40 & 22.79 & 22.84 & 3.74 & 2.28 & 32.97 & 23.84
        & 64.33 & 56.43 & 60.38 \\

        \rowcolor{blue!10}
        \method{\textbf{RASFT (Ours)}}
        & \textbf{66.39} & \textbf{23.59} & \textbf{27.10} & \textbf{7.92} & 3.96 & \textbf{37.34} & \textbf{27.72}
        & \textbf{65.05} & \textbf{56.50} & \textbf{60.78} \\

        \midrule

        & \multicolumn{7}{l}{\model{Qwen2.5-Math-7B}}
        & \multicolumn{3}{l}{\model{Qwen2.5-Coder-7B}} \\

        \method{Base}
        & 32.39 & 7.92 & 9.01 & 7.29 & 1.87 & 19.22 & 12.95
        & 20.39 & 41.59 & 30.99 \\

        \method{SFT}
        & 51.52 & 17.49 & 16.38 & 1.88 & 1.04 & 27.97 & 19.38
        & 66.12 & 51.29 & 58.71 \\

        \method{DFT~\cite{wu2025generalization}}
        & 68.28 & \textbf{32.96} & 32.06 & \textbf{8.97} & 4.80 & 45.31 & 32.06
        & 71.46 & 56.32 & 63.89 \\

        \method{ASFT~\cite{zhu2025anchored}}
        & 65.76 & 24.14 & 29.25 & \textbf{8.97} & 5.83 & 43.28 & 29.54
        & 69.28 & 53.29 & 61.29 \\

        \method{ProFiT~\cite{liu2026profit}}
        & 68.45 & 28.67 & 31.41 & 7.08 & 4.99 & 44.06 & 30.78
        & 71.57 & 55.44 & 63.51 \\

        \rowcolor{blue!10}
        \method{\textbf{RASFT (Ours)}}
        & \textbf{70.65} & 29.97 & \textbf{32.99} & 8.76 & \textbf{5.84} & \textbf{45.78} & \textbf{32.33}
        & \textbf{72.29} & \textbf{58.33} & \textbf{65.31} \\

        \midrule

        & \multicolumn{7}{l}{\model{Llama-3.2-3B}}
        & \multicolumn{3}{l}{\model{Llama-3.2-3B}} \\

        \method{Base}
        & 1.51 & 1.15 & 0.86 & 0.00 & \textbf{0.41} & 1.56 & 0.92
        & 8.35 & 9.54 & 8.95 \\

        \method{SFT}
        & 4.86 & 1.58 & 1.24 & 0.00 & 0.00 & 1.09 & 1.46
        & 19.74 & 23.65 & 21.70 \\

        \method{DFT~\cite{wu2025generalization}}
        & 7.71 & 3.07 & 2.03 & \textbf{0.21} & 0.21 & 2.19 & 2.57
        & 21.80 & 26.49 & 24.15 \\

        \method{ASFT~\cite{zhu2025anchored}}
        & 4.35 & 1.99 & 1.42 & \textbf{0.21} & 0.21 & 2.66 & 1.81
        & 20.73 & 29.13 & 24.93 \\

        \method{ProFiT~\cite{liu2026profit}}
        & \textbf{8.51} & 3.07 & \textbf{2.26} & \textbf{0.21} & 0.21 & 4.38 & 3.11
        & 17.95 & 25.98 & 21.97 \\

        \rowcolor{blue!10}
        \method{\textbf{RASFT (Ours)}}
        & 8.26 & \textbf{3.99} & 1.98 & \textbf{0.21} & 0.21 & \textbf{4.84} & \textbf{3.25}
        & \textbf{28.39} & \textbf{34.87} & \textbf{31.63} \\

        \bottomrule
    \end{tabular}
    \end{adjustbox}
    \endgroup

    \caption{\textbf{Performance comparison of SFT and its variants on mathematical and code benchmarks.} The left block reports mathematical reasoning results, while the right block reports code reasoning results. \textbf{Bold} numbers indicate the best performance within each model group. Blue-shaded rows denote our RASFT method.}
    \label{tab:math-code-results}
\end{table*}

\section{Experiments}
In this section, we evaluate RASFT to answer the following research questions (RQs): \textbf{RQ1}: How does RASFT compare with recent SFT variants on mathematical and code reasoning tasks?  \textbf{RQ2}: Is difficulty-adaptive weighting beneficial for balancing expert trajectories and model-generated rollouts?  \textbf{RQ3}: Do on-policy rollouts improve the training?   \textbf{RQ4}: How much does the inverse ratio contribute to performance?  \textbf{RQ5}: Can RASFT be effectively combined with different supervised fine-tuning objectives?

\subsection{Experimental Setup}
\paragraph{Models and Data.} We conduct fine-tuning experiments on two widely adopted model families, namely LLaMA~\cite{meta2024llama32} and Qwen2.5~\cite{qwen2025qwen25}. (i) \textbf{For mathematical reasoning experiments}, we select Qwen2.5-Math-1.5B, Qwen2.5-Math-7B~\cite{yang2024qwen25math}, and Llama-3.2-3B due to their strong mathematical capabilities. Following the experimental protocols of DFT~\cite{wu2025generalization} and ASFT~\cite{zhu2025anchored}, we use 10k samples from NuminaMath CoT~\cite{numina_math_datasets} for training. We evaluate mathematical reasoning performance on six benchmarks: 
MATH-500~\cite{lightman2023lets}, 
Minerva Math~\cite{lewkowycz2022solving}, 
OlympiadBench~\cite{he2024olympiadbench}, 
AIME 2024~\cite{maa2024aime}, 
AIME 2025~\cite{maa2025aime}, 
and AMC 2023~\cite{maa2023amc}. (ii) \textbf{For code reasoning experiments}, we select Qwen2.5-Coder-3B, Qwen2.5-Coder-7B~\cite{hui2024qwen25coder}, and Llama-3.2-3B, and use 10k samples from KodCode-V1-SFT-R1~\cite{xu2025kodcode} as the training dataset. We evaluate code reasoning performance on HumanEval~\cite{chen2021evaluating} and MBPP~\cite{austin2021program}. 

\paragraph{Baselines.} 
We compare our method with standard SFT, and several recent variants of SFT-based post-training methods, including: 
(i) SFT, a classical post-training paradigm that directly learns from expert demonstrations; 
(ii) DFT~\cite{wu2025generalization}, which dynamically rescales the token-level SFT objective according to token probabilities to improve generalization; 
(iii) ASFT~\cite{zhu2025anchored}, which augments DFT with a lightweight KL-based anchoring term to stabilize training and mitigate distributional drift; 
and (iv) ProFiT~\cite{liu2026profit}, which leverages probability-guided token selection to mask low-probability tokens and reduce overfitting to non-essential surface expressions.

\paragraph{Implementation Details.} For training, all models are optimized with a learning rate of $5 \times 10^{-5}$, a batch size of 256, a maximum input length of 2048 tokens, and 4 warm-up steps. For evaluation, we follow the evaluation protocols of two ICLR 2026 works, DFT~\cite{wu2025generalization} and ASFT~\cite{zhu2025anchored} to report the mean accuracy over 16 decoding runs for all models, using a temperature of 1.0 and a maximum generation length of 4096 tokens. See Appendix~\ref{app:training_evaluation_details} for details.

\subsection{Main Results Analysis (RQ1)}

\noindent \textbf{RASFT establishes superior performance across mathematical and code reasoning benchmarks.} As shown in Table~\ref{tab:math-code-results}, RASFT achieves the best average performance across all models on both mathematical and code reasoning tasks. On math benchmarks, it improves the strongest baselines by +2.72, +0.27, and +0.14 average points on Qwen2.5-Math-1.5B, Qwen2.5-Math-7B, and Llama-3.2-3B, respectively; on code benchmarks, the corresponding gains are +0.40, +1.42, and +6.70 points. These results indicate that rollout-adaptive supervision provides benefits beyond static expert imitation: verified rollouts supply additional correct reasoning trajectories, while the solvability-based weighting dynamically adjusts the reliance on expert demonstrations. The gains are especially clear on benchmarks such as OlympiadBench and AIME24 for Qwen2.5-Math-1.5B, where RASFT improves the best baseline scores from 24.11 to 27.10 and from 4.79 to 7.92, respectively.

\noindent \textbf{The gains of RASFT vary across model scales and task domains.} On mathematical reasoning, the improvement over the strongest baseline becomes smaller as model capacity increases:  on the MATH benchmark, the gain decreases from +5.50 points on Qwen2.5-Math-1.5B 
(66.39 vs. 60.89 for ASFT) to +2.20 points on Qwen2.5-Math-7B (70.65 vs. 68.45 for ProFiT). 
The same pattern appears in the average mathematical score, where the gain decreases from +2.72 to +0.27 points. This suggests that RASFT is especially helpful when the base model's reasoning distribution is under-optimized. For code reasoning, however, the scaling trend is less monotonic: RASFT improves the average score by +0.40 points on Qwen2.5-Coder-3B  and by +1.42 points on Qwen2.5-Coder-7B. Thus, the effectiveness of RASFT depends on both model capacity and task domain. Importantly, RASFT achieves the best average performance for both 7B mathematical and code models, indicating that its rollout-adaptive weighting remains effective for stronger pretrained policies.

\begin{table*}[t]
\centering
\caption{\textbf{Ablation Studies on Qwen2.5-Math-1.5B.} \textbf{Bold} numbers indicate the best performance within each block. $\uparrow$ and $\downarrow$ indicate performance gain or drop relative to the baseline configuration within each block respectively.}
\label{tab:master_ablation}

\small
\setlength{\tabcolsep}{8pt} 
\renewcommand{\arraystretch}{1.15}

\scalebox{0.9}{\begin{tabular}{llccccccc@{\hspace{15pt}}}
\toprule
\textbf{Ablation} & \textbf{Experiment} & \textbf{Math} & \textbf{Minerva} & \textbf{Olympiad} & \textbf{AIME24} & \textbf{AIME25} & \textbf{AMC23} & \textbf{Avg.} \\
\midrule

\multirow{4}{*}

\multirow{2}{*}{\ding{172} \textit{Difficulty Adaptivity}}
& w/o Adaptivity
& 63.04 & 21.24 & 23.61 & 5.22 & 3.95 & 36.09 & 25.53 \\
& w/ Adaptivity
& \textbf{66.39}\rlap{$_{\textcolor{orange}{\uparrow 3.35}}$}
& \textbf{23.59}\rlap{$_{\textcolor{orange}{\uparrow 2.35}}$}
& \textbf{27.10}\rlap{$_{\textcolor{orange}{\uparrow 3.49}}$}
& \textbf{7.92}\rlap{$_{\textcolor{orange}{\uparrow 2.70}}$}
& \textbf{3.96}\rlap{$_{\textcolor{orange}{\uparrow 0.01}}$}
& \textbf{37.34}\rlap{$_{\textcolor{orange}{\uparrow 1.25}}$}
& \textbf{27.72}\rlap{$_{\textcolor{orange}{\uparrow 2.19}}$} \\
\midrule

\multirow{2}{*}
{\ding{173} \textit{Exploration Strategy}}
& w/o Rollout
& 63.65 & 20.31 & 23.23 & 6.06 & 3.74 & 33.28 & 25.05 \\
& w/ Rollout
& \textbf{66.39}\rlap{$_{\textcolor{orange}{\uparrow 2.74}}$}
& \textbf{23.59}\rlap{$_{\textcolor{orange}{\uparrow 3.28}}$}
& \textbf{27.10}\rlap{$_{\textcolor{orange}{\uparrow 3.87}}$}
& \textbf{7.92}\rlap{$_{\textcolor{orange}{\uparrow 1.86}}$}
& \textbf{3.96}\rlap{$_{\textcolor{orange}{\uparrow 0.22}}$}
& \textbf{37.34}\rlap{$_{\textcolor{orange}{\uparrow 4.06}}$}
& \textbf{27.72}\rlap{$_{\textcolor{orange}{\uparrow 2.67}}$} \\
\midrule

\multirow{2}{*}{\ding{174} \textit{Policy Regularizer}}
& w/o Ratio
& 64.24 & 22.53 & 25.91 & \textbf{8.54} & \textbf{5.41} & 37.19 & 27.30 \\
& w/ Ratio
& \textbf{66.39}\rlap{$_{\textcolor{orange}{\uparrow 2.15}}$}
& \textbf{23.59}\rlap{$_{\textcolor{orange}{\uparrow 1.06}}$}
& \textbf{27.10}\rlap{$_{\textcolor{orange}{\uparrow 1.19}}$}
& 7.92\rlap{$_{\textcolor{teal}{\downarrow 0.62}}$}
& 3.96\rlap{$_{\textcolor{teal}{\downarrow 1.45}}$}
& \textbf{37.34}\rlap{$_{\textcolor{orange}{\uparrow 0.15}}$}
& \textbf{27.72}\rlap{$_{\textcolor{orange}{\uparrow 0.42}}$} \\

\midrule

\multirow{4}{*}{\ding{175} \textit{Transferability}}
& SFT
& 40.69 & 11.27 & 10.28 & 1.66 & 0.41 & 16.56 & 13.48 \\
& ~+ RASFT
& \textbf{42.96}\rlap{$_{\textcolor{orange}{\uparrow 2.27}}$}
& \textbf{12.91}\rlap{$_{\textcolor{orange}{\uparrow 1.64}}$}
& \textbf{11.26}\rlap{$_{\textcolor{orange}{\uparrow 0.98}}$}
& 0.83\rlap{$_{\textcolor{teal}{\downarrow 0.83}}$}
& 0.21\rlap{$_{\textcolor{teal}{\downarrow 0.20}}$}
& \textbf{18.12}\rlap{$_{\textcolor{orange}{\uparrow 1.56}}$}
& \textbf{14.38}\rlap{$_{\textcolor{orange}{\uparrow 0.90}}$} \\
& ProFiT
& 58.40 & 22.79 & 22.84 & 3.74 & 2.28 & 32.97 & 23.84 \\
& ~+ RASFT
& \textbf{62.99}\rlap{$_{\textcolor{orange}{\uparrow 4.59}}$}
& \textbf{25.99}\rlap{$_{\textcolor{orange}{\uparrow 3.20}}$}
& \textbf{26.25}\rlap{$_{\textcolor{orange}{\uparrow 3.41}}$}
& \textbf{5.01}\rlap{$_{\textcolor{orange}{\uparrow 1.27}}$}
& \textbf{4.17}\rlap{$_{\textcolor{orange}{\uparrow 1.89}}$}
& \textbf{36.09}\rlap{$_{\textcolor{orange}{\uparrow 3.12}}$}
& \textbf{26.75}\rlap{$_{\textcolor{orange}{\uparrow 2.91}}$} \\
\bottomrule
\end{tabular}}
\end{table*}

\subsection{Ablation Study (RQ2-5)}
To evaluate the effectiveness of RASFT, we conduct ablation studies on Qwen2.5-Math-1.5B and report results on Math benchmarks, covering the following components:  \textit{\ding{172} Difficulty-adaptive training}, \textit{\ding{173} The on-policy rollout},  \textit{\ding{174} The inverse importance sampling ratio} and \textit{\ding{175} The transferability of RASFT to SFT and variants algorithms}.

\textbf{\textit{\ding{172} Difficulty-adaptive training.}}
To examine whether RASFT effectively assigns different weights to trajectories within each group according to problem difficulty, we introduce a variant that assigns equal weights to the expert trajectory and verified rollout trajectories for all problems, regardless of their estimated difficulty. As shown in Table~\ref{tab:master_ablation}, removing difficulty adaptivity leads to a clear performance drop, with the average score decreasing from $27.72$ to $25.53$. The improvement brought by adaptive weighting is consistent across all six benchmarks: $+3.35$ on Math, $+2.35$ on Minerva Math, $+3.49$ on OlympiadBench, $+2.70$ on AIME24, $+0.01$ on AIME25, and $+1.25$ on AMC23. The gains are especially pronounced on Math, OlympiadBench, and AIME24, indicating that treating all problems with the same expert-rollout balance is suboptimal.

\textbf{\textit{\ding{173} The on-policy rollout.}} To evaluate the significance of online exploration and dynamic solvability estimation, we remove the on-policy rollout sampling mechanism and assign a fixed static weight to the expert trajectories. As summarized in Table~\ref{tab:master_ablation}, discarding the on-policy rollouts triggers an across-the-board performance degradation on every single evaluated benchmark, with the average score decreasing from 27.72 to 25.05. The drop is particularly pronounced on challenging datasets such as OlympiadBench (-3.87\%) and AMC23 (-4.06\%). This uniform regression demonstrates that pairing external expert trajectories with the model's own attempts in a local comparison group is vital for effective post-training. Without this online feedback loop, the optimization lapses into rigid imitation of offline data without online adaptive calibration, failing to calibrate the training pressure according to the model's evolving capacity and thereby limiting the activation of its latent reasoning pathways.

\textbf{\textit{\ding{174} The inverse importance sampling ratio.}} To examine the impact of the sequence-level inverse importance sampling ratio $r_{i,j}(\theta)$, we evaluate a variant by setting $r_{i,j}(\theta)$ uniformly, which removes the reference-relative policy regularization. As shown in Table~\ref{tab:master_ablation}, introducing the inverse ratio improves the overall performance from 27.30\% to 27.72\%, yielding advancements on primary benchmarks such as Math (+2.15\%), Minerva Math (+1.06\%), and OlympiadBench (+1.19\%). This growth confirms that the sequence-level ratio effectively modulates the update magnitude by dampening gradient updates when the current policy already aligns closely with the expert trajectories. By preventing the model from overfitting to offline demonstrations, this regularization mechanism safeguards the model's pretrained reasoning priors and ensures more balanced generalization across the  mathematical benchmarks.

\textbf{\textit{\ding{175} The transferability of RASFT to SFT and its variants.}}
To evaluate whether RASFT can generalize across different SFT-style objectives, we integrate it with standard SFT and ProFiT. For the SFT variant, we remove the token-level probability factor $p_{i,j,t}^{\theta}$ from Eq.~\ref{eq:weight}. For the ProFiT variant, we additionally apply its token masking strategy by filtering tokens whose predictive probabilities are below 0.1. Since Table~\ref{tab:math-code-results} reports the default RASFT built upon DFT, this ablation compares RASFT when combined with SFT, ProFiT, and DFT. As shown in Tables~\ref{tab:math-code-results} and~\ref{tab:master_ablation}, RASFT consistently improves all three SFT-style baselines, confirming its transferability. In particular, it improves the average score by +0.90 points over SFT, +2.91 points over ProFiT, and +3.53 points over DFT. The gains are also evident on difficult competition benchmarks: RASFT raises ProFiT on AIME25 from 2.28\% to 4.17\%, and improves DFT on OlympiadBench from 22.42\% to 27.10\%. These results support our motivation that difficult problems benefit from stronger expert guidance, while easier problems can rely more on verified model-generated trajectories. Moreover, the larger gains on ProFiT and DFT suggest that RASFT is complementary to token-level rescaling or filtering methods, as it introduces sequence-level rollout adaptivity and candidate-pool weighting.

\section{Analysis and Discussion}

\subsection{Comparison with RL methods}
To contextualize RASFT within the landscape of recent RL alignment paradigms, we evaluate our method against two representative RL-driven algorithms: (i) GRPO~\cite{shao2024deepseekmath}, a prominent pure on-policy reinforcement learning baseline, and (ii) LUFFY~\cite{yan2025luffy}, a mixed-policy framework that augments on-policy exploration with off-policy reasoning guidance via regularized importance sampling. For all 3 experiments, we train on Qwen2.5-Math-1.5B. More training details are provided in Appendix~\ref{app:rl_baseline_details}.

\begin{figure}[t]
    \centering
    \includegraphics[width=1\linewidth,height=0.25\textheight]
    {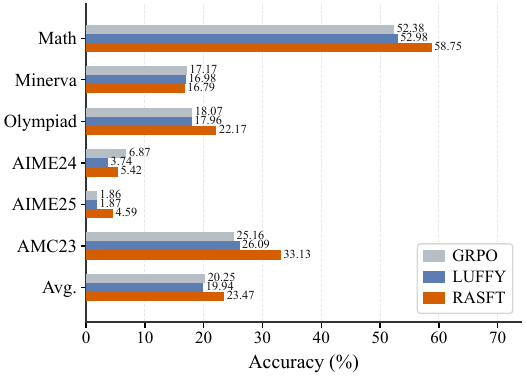}
    \caption{Comparison between RASFT and representative RL methods (GRPO and LUFFY). RASFT demonstrates superior performance by dynamically balancing offline expert guidance and online exploration. More training details are provided in Appendix~\ref{app:rl_baseline_details}.}
    \label{fig:rl_comparison}
    \vspace{-1em}
\end{figure}

As illustrated in Figure~\ref{fig:rl_comparison}, RASFT secures the highest average accuracy of 23.47\%. This robust performance is characterized by two primary insights: first, our solvability-aware modulation adaptively scales up the expert trajectory weight when online exploration fails, which successfully bypasses the exploration cold-start and optimization stagnation that cause pure on-policy methods like GRPO to collapse on ultra-hard tasks like AIME25; second, while unconstrained exploration enables GRPO to occasionally discover highly alternative reasoning paths on specific domains like Minerva Math, RASFT maintains a superior and more stable overall optimization trajectory by utilizing the sequence-level inverse importance sampling ratio to dampen redundant updates and safeguard pretrained reasoning priors.

\subsection{Training Dynamics Analysis}

\begin{figure}[t]
  \centering
   \includegraphics[width=0.85\linewidth,height=0.25\textheight]{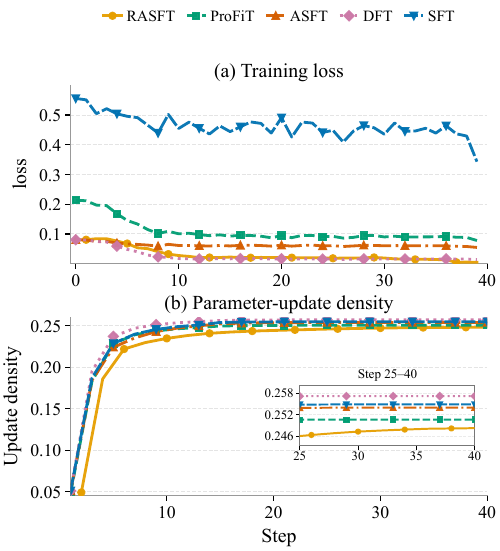}

\caption{
\textbf{Training dynamics.}
Training loss and parameter-update density across SFT variants; density is the fraction of updates with magnitude above $10^{-5}$.
}

    \label{fig:training_dynamics}
    \vspace{-0.9em}
\end{figure}

We further analyze the training dynamics of different SFT-style methods by tracking the training objective and parameter-update magnitude. 
For the latter, we define parameter-update density as the fraction of parameters whose absolute update magnitude exceeds $10^{-5}$ at each optimization step, which reflects how broadly model parameters are modified during training. As shown in Figure~\ref{fig:training_dynamics}, RASFT maintains a stable training trajectory. 
Its training objective decreases smoothly and does not exhibit noticeable oscillation, showing a stability pattern consistent with strong SFT variants such as DFT, ASFT, and ProFiT. 
This indicates that introducing verified rollouts and difficulty-adaptive weighting does not destabilize supervised post-training. 
Meanwhile, the parameter-update density of all methods increases rapidly in the early stage and then saturates, while RASFT remains comparable to or slightly lower than the other methods in the late stage. 
Specifically, RASFT reaches a final update density of $0.2483$, compared with $0.2507$ for ProFiT, $0.2540$ for ASFT, $0.2573$ for DFT, and $0.2549$ for SFT; the last-five-step average follows the same pattern, with RASFT obtaining the lowest density of $0.2480$.  This supports our motivation that rollout-adaptive weighting, together with the inverse ratio, provides targeted supervision while better preserving the pretrained reasoning prior.

\subsection{Sensitivity to the Rollout Number}

\begin{figure}[t]
    \centering
    \includegraphics[width=0.8\linewidth,height=0.13\textheight]
    {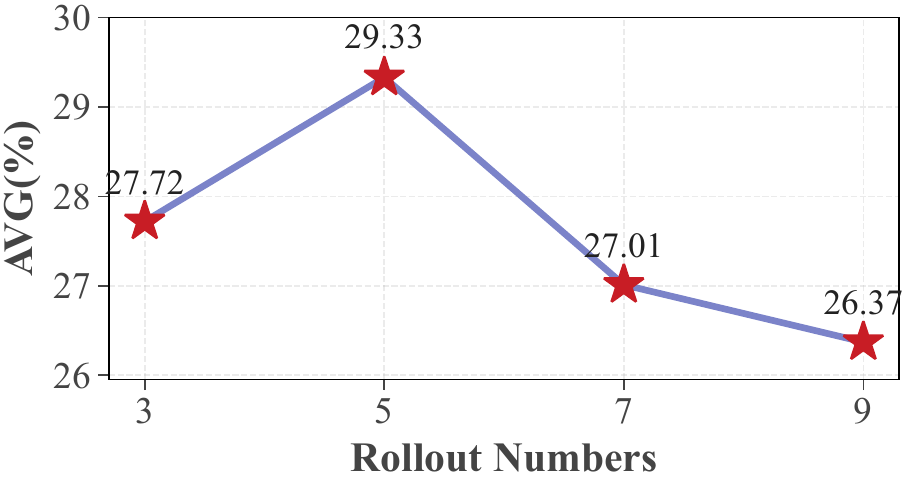}
    \caption{\textbf{Sensitivity to the Rollout Number.}}
    \label{fig:rollout_sensitivity}
    \vspace{-0.9em}
\end{figure}

To examine the sensitivity of RASFT to the rollout number, we vary $K$ in $\{3,5,7,9\}$ and report the average mathematical reasoning accuracy in Figure~\ref{fig:rollout_sensitivity}. The performance first increases from 27.72 at $K=3$ to 29.33 at $K=5$, yielding a gain of 1.61 points, but then drops to 27.01 at $K=7$ and 26.37 at $K=9$, corresponding to decreases of 2.32 and 2.96 points compared with $K=5$. This non-monotonic trend shows that simply increasing the rollout budget does not necessarily improve RASFT: a small rollout number may provide insufficient on-policy feedback, while excessive rollouts may introduce redundant self-generated trajectories and weaken the relative contribution of expert demonstrations after candidate-pool normalization. Nevertheless, all rollout settings still outperform the other baselines in Table~\ref{tab:math-code-results}; even the weakest setting, K=9, achieves 26.37 average accuracy, which remains higher than the strongest baseline average of 25.00. The complete statistical results are provided in Appendix~\ref{app:Sensitivity to Rollout Number}.

\section{Related Work}
\paragraph{Objective Refinement of SFT.}
Supervised SFT with Chain-of-Thought (CoT) demonstrations is widely used to activate the reasoning abilities of large language models~\cite{wei2022chain,chung2022scaling,mukherjee2023orca,yue2023mammoth}. However, standard cross-entropy training treats offline demonstrations as fixed targets and uniformly maximizes the likelihood of the reference response, which can lead to overfitting to demonstration-specific patterns and limited generalization. 
Recent methods improve SFT by reweighting or selecting supervision signals from offline data. 
DFT~\cite{wu2025generalization} rescales the objective according to the model's prediction probability, ASFT~\cite{zhu2025anchored} adds reference anchoring to stabilize this reweighting, and ProFiT~\cite{liu2026profit} selects high-value tokens to reduce unnecessary fitting. 
However, they still center on the expert response and overlook model-generated reasoning trajectories.

\paragraph{Rollout-Based Policy Optimization for Reasoning.}
Recent post-training methods for reasoning increasingly rely on reinforcement learning with sampled trajectories and outcome feedback~\cite{cobbe2021training,yuan2023scaling,zelikman2022star,gulcehre2023reinforced,deepseekai2025deepseekr1}. PPO~\cite{schulman2017proximal} has been widely used as a policy-optimization algorithm, using clipped policy updates to improve training stability. Building on this direction, GRPO~\cite{shao2024deepseekmath} optimizes relative advantages within sampled groups, emphasizing the importance of comparing multiple candidate reasoning trajectories rather than fitting one reference response. More recently, LUFFY~\cite{yan2025luffy} incorporates off-policy reasoning guidance into RL training, showing the value of combining online exploration with external trajectory information. However, these RL methods require explicit policy optimization and extensive rollouts, whereas RASFT uses verified rollouts as adaptive supervision within an SFT objective.

\section{Conclusions and Future Work}
We introduce RASFT, a rollout-adaptive supervised fine-tuning method that calibrates expert supervision using the model's own rollout behavior. Empirical results show that RASFT improves over SFT and SFT variants across multiple models and challenging mathematical and code reasoning tasks. Compared with RL methods, RASFT also achieves stronger overall performance. Overall, our results suggest that expert demonstrations can serve as adaptive guidance rather than fixed imitation targets, while the inverse policy ratio helps preserve useful reasoning priors. Looking forward, a promising direction is to extend RASFT to open-ended reasoning tasks where reliable automatic correctness signals are harder to define. 
Another direction is to study how rollout-adaptive supervision scales with stronger models and larger training corpora across broader domains, especially when combined with more diverse verifiers and broader forms of verified, diverse self-generated supervision.

\clearpage
\section*{Limitations}
RASFT introduces additional computational cost compared with standard SFT, since it requires periodically generating on-policy rollouts and verifying their correctness. Although we reduce this cost by applying rollouts only to selected training instances, the overall training pipeline is still more expensive than methods that only optimize offline demonstrations. This also makes the method sensitive to the rollout budget: using too few rollouts may provide insufficient feedback, while excessive rollouts can increase computation and introduce redundant verified trajectories. Our method also relies on task-specific verification signals. In mathematical reasoning, we verify sampled responses by extracting the final answer from $\backslash\texttt{boxed}\{\}$ and comparing it with the ground-truth answer, following the common practice of evaluating mathematical problem solving through final-answer correctness~\cite{hendrycks2021math}. In code reasoning, we use assertion-based tests, following the common practice of evaluating generated programs with executable test cases~\cite{hendrycks2021measuring,li2022competition}. Therefore, RASFT is most directly applicable to domains where reliable automatic verification is available. Extending the method to open-ended reasoning tasks without clear correctness signals remains challenging. Such tasks often rely on human preferences or learned reward models, which require costly annotation and may depend on task-specific preference modeling~\cite{stiennon2020learning,bai2022training}. Recent LLM-based judges provide a scalable alternative, but they may still introduce evaluator-specific biases, including positional or instruction-following failures~\cite{zheng2023judging,liu2023geval,wang2024large,zeng2023evaluating}.

\bibliography{custom}

\clearpage
\appendix

\section{Training and Evaluation Details}
\label{app:training_evaluation_details}
\paragraph{Training details.}
For all experiments, we train all models with AdamW and a cosine learning-rate schedule. We use a learning rate of $5\times10^{-5}$, an effective batch size of 256, a batch size per-device of 2, a maximum input length of 2048 tokens, 4 warm-up steps, bf16 precision, gradient checkpointing and gradient clipping with a maximum norm of 1.0. All models are trained for 1 epoch with random seed 42. For RASFT, we generate 3 rollouts per selected prompt with temperature 0.9, top-p 0.95, repetition penalty 1.1, and a maximum rollout length of 1024 tokens. We clip the ratio to [0.4, 1.1]. 

For RASFT on mathematical tasks, we set $W_e^{\min}=W_r=1.0$ and $W_e^{\max}=3.0$ for the two Qwen2.5-Math models, and $W_e^{\min}=W_r=1.5$ and $W_e^{\max}=3.0$ for Llama-3.2-3B. The correctness of sampled responses is verified by extracting and comparing the final answer enclosed in the last $\backslash\texttt{boxed}\{\}$ in the response. For code tasks, we set $W_e^{\min}=1.0$, $W_r=1.5$, and $W_e^{\max}=1.8$  for the two Qwen2.5-Coder models, and $W_e^{\min}=W_r=1.5$ and $W_e^{\max}=3.0$ for Llama-3.2-3B. The correctness of generated code samples is verified using assertion-based tests.

\paragraph{Evaluation details.}
 We follow the evaluation protocols of DFT~\cite{wu2025generalization} and ASFT~\cite{zhu2025anchored}. For mathematical reasoning, we evaluate all methods on the 6 mathematical reasoning benchmarks mentioned above. For each problem, we sample 16 responses with temperature 1.0 and a maximum generation length of 4096 tokens. We report the average accuracy over the 16 sampled responses. Correctness is determined by extracting the final answer and comparing it with the ground-truth answer; for code reasoning, we evaluate all methods on HumanEval~\cite{chen2021evaluating} and MBPP~\cite{austin2021program}. We use the same decoding setting as in mathematical reasoning: temperature 1.0, 16 sampled responses per problem, and a maximum generation length of 4096 tokens. We report the mean accuracy over the 16 sampled responses, where the correctness is determined by executing the generated code against assertion-based tests.

\paragraph{Computation cost.}
We report the computational cost of different training modes in
Table~\ref{tab:compute_cost}. For the baseline methods reported in
Table~\ref{tab:math-code-results}, SFT, DFT, and ProFiT are expected to
have nearly identical computation costs, since DFT and ProFiT do not introduce
additional model components and only modify the token-level training objective.
In contrast, ASFT introduces an additional reference model, leading to higher
memory usage and different computational efficiency. Therefore, we report the training costs of SFT, ASFT, and RASFT on the math task using 4× A100 SXM4 GPUs.
\begin{table}[t]
\centering
\footnotesize
\setlength{\tabcolsep}{4.2pt}
\renewcommand{\arraystretch}{1.12}
\begin{tabular}{@{}llccc@{}}
\toprule
\textbf{Model} & \textbf{Mode} & \textbf{TFLOPs/h} & \textbf{Mem (GB)} & \textbf{Hours} \\
\midrule
\multirow{3}{*}{Qwen2.5-Math-1.5B}
& SFT   & $2.40{\times}10^{5}$ & 81.37  & 0.227 \\
& ASFT  & $1.70{\times}10^{5}$ & 117.69 & 0.320 \\
& RASFT & $2.61{\times}10^{5}$ & 154.40 & 0.554 \\
\midrule
\multirow{3}{*}{Qwen2.5-Math-7B}
& SFT   & $5.38{\times}10^{5}$ & 145.76 & 0.545 \\
& ASFT  & $3.95{\times}10^{5}$ & 217.15 & 0.743 \\
& RASFT & $4.95{\times}10^{5}$ & 211.11 & 1.331 \\
\bottomrule
\end{tabular}
\caption{Computation cost comparison across different training modes.}
\label{tab:compute_cost}
\end{table}

\begin{table}[t]
\centering
\begingroup
\small
\setlength{\tabcolsep}{4.0pt}
\renewcommand{\arraystretch}{1.06}

\begin{tabular}{@{}lcccc@{}}
\toprule
\textbf{Benchmark} & \textbf{$K=3$} & \textbf{$K=5$} & \textbf{$K=7$} & \textbf{$K=9$} \\
\midrule
MATH-500       & 66.39 & \textbf{68.34} & 66.08 & 65.78 \\
Minerva        & 23.59 & \textbf{25.13} & 21.63 & 23.59 \\
OlympiadBench  & 27.10 & \textbf{28.87} & 27.74 & 26.41 \\
AIME24         & 7.92  & \textbf{10.00} & 7.91  & 5.83  \\
AIME25         & 3.96  & \textbf{5.85}  & 3.54  & 2.09  \\
AMC23          & 37.34 & \textbf{37.81} & 35.15 & 34.53 \\
\midrule
Avg.           & 27.72 & \textbf{29.33} & 27.01 & 26.37 \\
\bottomrule
\end{tabular}

\endgroup
\caption{Detailed mathematical reasoning results of RASFT with different rollout numbers $K$ on Qwen2.5-Math-1.5B. We report accuracy on each benchmark and the average accuracy across all six benchmarks.}
\label{tab:rollout-number-sensitivity}
\end{table}


\section{Detailed Results for Rollout Number Sensitivity}
\label{app:Sensitivity to Rollout Number}

This appendix provides the complete benchmark-level results for the rollout number sensitivity analysis. 
We evaluate RASFT on Qwen2.5-Math-1.5B using 10k samples from NuminaMath CoT, varying the number of sampled rollouts $K$ in $\{3,5,7,9\}$. 
The results are reported in Table~\ref{tab:rollout-number-sensitivity}.  
Overall, the best average performance is obtained when $K=5$, suggesting that a moderate number of rollouts provides more informative on-policy feedback than a smaller rollout budget. 
However, further increasing $K$ to 7 or 9 does not lead to additional gains. 
This supports the observation that excessive rollouts may introduce redundant self-generated trajectories and reduce the relative influence of expert demonstrations after candidate-pool normalization. Although we use $K=3$ in the main experiments for efficiency, all evaluated rollout settings achieve higher average accuracy than the strongest non-RASFT baseline reported in Table~\ref{tab:math-code-results} for Qwen2.5-Math-1.5B.

\section{Training Details for RL Baselines}
\label{app:rl_baseline_details}

\begin{table}[t]
\centering
\small
\begin{tabular}{l c}
\hline
\textbf{Hyperparameter} & \textbf{Value} \\
\hline
Training batch size & 64 \\
Validation batch size & 128 \\
Maximum prompt length & 1024 \\
Maximum response length & 1024 \\
Actor learning rate & $1\times10^{-6}$ \\
Training rollout temperature & 1.0 \\
Validation rollout temperature & 0.6 \\
Number of training rollouts & 4 \\
Number of validation rollouts & 1 \\
Entropy coefficient & 0.001 \\
KL coefficient & 0.0 \\
Training epochs & 10 \\
Number of GPUs & 4 \\
\hline
\end{tabular}
\caption{Training hyperparameters used for GRPO and LUFFY. We follow the configuration provided by LUFFY.}
\label{tab:rl_training_details}
\end{table}

We compare RASFT with two representative RL-based training methods, GRPO~\cite{shao2024deepseekmath} and LUFFY~\cite{yan2025luffy}, on Qwen2.5-Math-1.5B. For RASFT, we use the same training configuration as described in Appendix~\ref{app:training_evaluation_details}. For GRPO and LUFFY, we follow the configuration provided by LUFFY. For all 3 experiments, we sample 2000 from  NuminaMath CoT as training data. The main hyperparameters are summarized in Table~\ref{tab:rl_training_details}.

For LUFFY, we keep its default mixed-policy training configuration. In particular, we enable the token-level off-policy loss, use no off-policy normalization, and adopt the \texttt{p\_div\_p\_0.1} off-policy reshaping strategy. For GRPO, we use the group-relative advantage estimator under the same rollout and optimization settings.

\section{Case Studies}
\label{app:case-studies}
\subsection{Case 1}
\label{app:case-rainfall}

\begin{questionbox}
It rained some inches on Monday. On Tuesday, it rained \(0.7\) inches less than on Monday, and it rained \(0.2\) inches on Tuesday. How much did it rain on Monday?
\end{questionbox}

\begin{standardbox}
Let \(M\) denote the amount of rain on Monday. Since Tuesday's rainfall was \(0.7\) inches less than Monday's rainfall, and Tuesday's rainfall was \(0.2\) inches, we have
\[
M - 0.7 = 0.2.
\]
Solving for \(M\),
\[
M = 0.2 + 0.7 = 0.9.
\]
Therefore, it rained
\[
\boxed{0.9}
\]
inches on Monday.
\end{standardbox}
\begin{rolloutonebox}
The model denotes Monday's rainfall as \(M\). Since Tuesday's rainfall is \(0.7\) inches less than Monday's rainfall, it sets up
\[
M - 0.7 = 0.2.
\]
Solving the equation gives
\[
M = 0.2 + 0.7 = 0.9.
\]
Thus, the model predicts
\[
\boxed{0.9}.
\]
\end{rolloutonebox}

\begin{rollouttwobox}
The model writes the relation between Monday's and Tuesday's rainfall as
\[
M - 0.7 = 0.2.
\]
It then solves for \(M\):
\[
M = 0.2 + 0.7 = 0.9.
\]
Therefore, the model predicts
\[
\boxed{0.9}.
\]
\end{rollouttwobox}

\begin{rolloutthreebox}
The model again represents Tuesday's rainfall as \(M - 0.7\), where \(M\) is Monday's rainfall. Since Tuesday's rainfall is \(0.2\), it obtains
\[
M - 0.7 = 0.2.
\]
Therefore,
\[
M = 0.9.
\]
Thus, the model predicts
\[
\boxed{0.9}.
\]
\end{rolloutthreebox}

\subsection{Case 2}
\label{app:case-binomial}

\begin{questionbox}
Let
\[
(x+2)^n = a_0 + a_1x + a_2x^2 + \cdots + a_nx^n,
\]
where \(n \in \mathbb{N}^*\), \(n \geq 2\), and \(a_0\), \(a_1\), \(a_2\) form an arithmetic sequence.

\smallskip
\noindent
(1) Find the middle term of the expansion of \((x+2)^n\).

\smallskip
\noindent
(2) Find the sum of the coefficients of the terms with odd powers of \(x\) in the expansion of \((x+2)^n\).
\end{questionbox}

\begin{standardbox}
The coefficient of \(x^r\) in \((x+2)^n\) is
\[
a_r = \binom{n}{r}2^{n-r}.
\]
Thus,
\[
a_0 = 2^n,\qquad
a_1 = n2^{n-1},\qquad
a_2 = \binom{n}{2}2^{n-2}.
\]
Since \(a_0,a_1,a_2\) form an arithmetic sequence,
\[
2a_1 = a_0 + a_2.
\]
Substituting the coefficients gives
\[
\begin{aligned}
2n2^{n-1}
&= 2^n + \binom{n}{2}2^{n-2}, \\
n^2 - 9n + 8 &= 0.
\end{aligned}
\]
Hence \(n=1\) or \(n=8\). Since \(n\geq 2\), we have \(n=8\).

For \(n=8\), the expansion has \(9\) terms, so the middle term is the fifth term:
\[
T_5 = \binom{8}{4}2^4x^4 = 1120x^4.
\]

For the sum of the coefficients of odd powers of \(x\), let
\[
f(x)=(x+2)^8.
\]
Then
\[
f(1)=3^8,\qquad f(-1)=1.
\]
Therefore,
\[
\begin{aligned}
a_1+a_3+a_5+a_7
&= \frac{f(1)-f(-1)}{2} \\
&= \frac{3^8-1}{2} \\
&= 3280.
\end{aligned}
\]
Thus, the answers are
\[
\boxed{1120x^4}
\qquad\text{and}\qquad
\boxed{3280}.
\]
\end{standardbox}

\begin{rolloutonebox}
The model correctly uses the arithmetic-sequence condition
\[
2a_1 = a_0 + a_2,
\]
where
\[
\begin{aligned}
a_0 &= 2^n, \\
a_1 &= \binom{n}{1}2^{n-1}, \\
a_2 &= \binom{n}{2}2^{n-2}.
\end{aligned}
\]
It obtains
\[
n^2 - 9n + 8 = 0,
\]
so \(n=8\). Therefore, the middle term is
\[
\binom{8}{4}2^4x^4 = 1120x^4.
\]

For the odd-power coefficients, the model evaluates
\[
\begin{aligned}
\frac{f(1)-f(-1)}{2}
&= \frac{3^8-1}{2} \\
&= 3280.
\end{aligned}
\]
Thus, this rollout predicts the correct answers:
\[
\boxed{1120x^4}
\qquad\text{and}\qquad
\boxed{3280}.
\]
\end{rolloutonebox}

\begin{rollouttwobox}
The model correctly derives
\[
n^2 - 9n + 8 = 0,
\]
and therefore correctly obtains \(n=8\). It also correctly identifies the middle term:
\[
\binom{8}{4}2^4x^4 = 1120x^4.
\]

However, when computing the sum of the odd-power coefficients directly, it writes
\[
\begin{aligned}
S
&= \binom{8}{1}2^7
 + \binom{8}{3}2^5 \\
&\quad + \binom{8}{5}2^3
 + \binom{8}{7}2.
\end{aligned}
\]
This expression is correct, but the model evaluates it incorrectly as
\[
S = 1056.
\]
The correct value should be
\[
S = 3280.
\]
Thus, this rollout has a local arithmetic error in the final computation and predicts
\[
\boxed{1056}.
\]
\end{rollouttwobox}

\begin{rolloutthreebox}
The model starts from the correct condition
\[
2a_1 = a_0 + a_2,
\]
but then incorrectly simplifies the resulting equation as
\[
2n = 2 + \frac{n(n-1)}{2},
\]
which leads to
\[
n=4.
\]
This is incorrect. The correct simplification gives
\[
n^2 - 9n + 8 = 0,
\]
so \(n=8\).

Because of this early algebraic error, the model gives the wrong middle term:
\[
24x^2.
\]

It then continues with \(n=4\) and computes
\[
f(1)-f(-1)=3^4-1=80.
\]
This also misses the required factor of \(1/2\) for extracting odd-power coefficients. Therefore, this rollout contains both an incorrect value of \(n\) and an incorrect odd-coefficient computation. It predicts
\[
\boxed{80}.
\]
\end{rolloutthreebox}

\subsection{Case 3}
\label{app:case-next-greater-element}

\begin{questionbox}
\begin{lstlisting}[language=Python]
def nextGreaterElement(arr):
    """
    Modify the array such that each element in the array is replaced with the next
    greater element present on its right side in the array. If there is no greater
    element, replace it with -1.
    
    Parameters:
    arr (List[int]): An array of integers
    
    Returns:
    List[int]: A new array with each element replaced by the next greater element
    
    Examples:
    >>> nextGreaterElement([4, 5, 2, 10, 8])
    [5, 10, 10, -1, -1]
    
    >>> nextGreaterElement([3, 7, 1, 7, 8, 4, 5])
    [7, 8, 7, 8, -1, 5, -1]
    """
\end{lstlisting}
\end{questionbox}

\begin{standardbox}
\begin{lstlisting}[language=Python]
def nextGreaterElement(arr):
    if not arr:
        return []

    n = len(arr)
    res = [-1] * n
    stack = []

    for i in range(n - 1, -1, -1):
        current = arr[i]

        while stack and stack[-1] <= current:
            stack.pop()

        if stack:
            res[i] = stack[-1]

        stack.append(current)

    return res
\end{lstlisting}
\end{standardbox}

\begin{rolloutonebox}
\begin{lstlisting}[language=Python]
def nextGreaterElement(arr):
    stack = []
    result = [-1] * len(arr)

    for i in range(len(arr)):
        while stack and arr[i] > arr[stack[-1]]:
            j = stack.pop()
            result[j] = arr[i]

        stack.append(i)

    return result
\end{lstlisting}
\end{rolloutonebox}

\begin{rollouttwobox}
\begin{lstlisting}[language=Python]
def nextGreaterElement(arr):
    stack = []
    result = []

    for num in reversed(arr):
        while stack and stack[-1] <= num:
            stack.pop()

        if stack:
            result.append(stack[-1])
        else:
            result.append(-1)

        stack.append(num)

    result.reverse()
    return result
\end{lstlisting}
\end{rollouttwobox}

\begin{rolloutthreebox}
\begin{lstlisting}[language=Python]
def nextGreaterElement(arr):
    stack = []
    result = []

    for num in reversed(arr):
        while stack and stack[-1] <= num:
            stack.pop()

        if stack:
            result.append(stack[-1])
        else:
            result.append(-1)

        stack.append(num)

    result.reverse()
    return result
\end{lstlisting}
\end{rolloutthreebox}

\subsection{Case 4}
\label{app:case-pair-participants}

\begin{questionbox}
\begin{lstlisting}[language=Python]
def pair_participants(n, skill_levels):
    """
    Pairs participants such that the difference in their skill levels is minimized.
    If the number of participants is odd, one participant will be left unpaired.

    Parameters:
    n (int): Number of participants.
    skill_levels (list): List of skill levels of participants in non-decreasing order.

    Returns:
    list: A list of tuples where each tuple represents a pair of participants.
    If there is an odd number of participants, the last element will be a single skill level.

    Examples:
    >>> pair_participants(6, [1, 2, 3, 5, 9, 12])
    [(1, 2), (3, 5), (9, 12)]

    >>> pair_participants(5, [1, 1, 3, 4, 9])
    [(1, 1), (3, 4), (9,)]
    """
    # Implementation here
\end{lstlisting}
\end{questionbox}

\begin{standardbox}
\begin{lstlisting}[language=Python]
def pair_participants(n, skill_levels):
    pairs = []

    for i in range(0, n, 2):
        if i + 1 < n:
            pairs.append((
                skill_levels[i],
                skill_levels[i + 1]
            ))
        else:
            pairs.append((skill_levels[i],))

    return pairs
\end{lstlisting}
\end{standardbox}

\begin{rolloutonebox}
\begin{lstlisting}[language=Python]
def pair_participants(n, skill_levels):
    if n % 2 == 0:
        pairs = []

        for i in range(0, n, 2):
            pairs.append((
                skill_levels[i],
                skill_levels[i + 1]
            ))

        return pairs
    else:
        pairs = []

        for i in range(0, n - 1, 2):
            pairs.append((
                skill_levels[i],
                skill_levels[i + 1]
            ))

        pairs.append(skill_levels[-1])
        return pairs
\end{lstlisting}
\end{rolloutonebox}

\begin{rollouttwobox}
\begin{lstlisting}[language=Python]
def pair_participants(n, skill_levels):
    if n % 2 == 0:
        pairs = []

        for i in range(0, n, 2):
            pairs.append((
                skill_levels[i],
                skill_levels[i + 1]
            ))

        return pairs
    else:
        pairs = []

        for i in range(0, n - 1, 2):
            pairs.append((
                skill_levels[i],
                skill_levels[i + 1]
            ))

        pairs.append(skill_levels[-1])
        return pairs
\end{lstlisting}
\end{rollouttwobox}

\begin{rolloutthreebox}
\begin{lstlisting}[language=Python]
def pair_participants(n, skill_levels):
    if n % 2 == 0:
        return list(zip(
            skill_levels[::2],
            skill_levels[1::2]
        ))
    else:
        return (
            list(zip(
                skill_levels[::2],
                skill_levels[1::2]
            ))
            + [(skill_levels[-1],)]
        )
\end{lstlisting}
\end{rolloutthreebox}

\end{document}